\documentclass[sigconf, nonacm]{acmart}
\usepackage{makecell}
\usepackage{multirow}
\usepackage{algorithm}
\usepackage{algorithmic}
\usepackage{graphicx}
\usepackage{xurl}
\AtBeginDocument{%
  }

\setcopyright{none}
\renewcommand\footnotetextcopyrightpermission[1]{}
\makeatletter\renewcommand{\ALC@linenodelimiter}{.}\makeatother

\begin{document}

\title{Autonomous Multi-modal Agents for ITOps Observability}

\author{Gunjan Siddharth, Sanjana Garai, Utsav Shekhar, Prakhar Jain, Sarvam Maheshwari, Balaji Ganesan, Avirup Saha, Seema Nagar, Pratik Mishra, Felix George, Suranjana Samanta, Harshit Kumar, Mudit Verma, Prateeti Mohapatra, Ravi Jagannathan, Amit Paradkar, Sameep Mehta, Pratibha Moogi}
\affiliation{\institution{IBM Software}\country{India}}
\email{pratibha.moogi@ibm.com}


\renewcommand{\shortauthors}{Siddharth et al.}

\begin{abstract}
ITOps observability products must integrate data from different tools and systems. We present ITOps Observability Studio, a multi-modal multi-agent system that orchestrates observability workflows through collaborative AI agents. Its three-tier architecture features inter-agent communication, specialized LLM-powered agents for data access, schema operations and visualization, and unified tool integration. The system supports natural language to GraphQL translation, GraphQL schema generation, schema mapping, trace data lifecycle management and multi-tool workflows for reliability engineering. Its agents are instrumented with agent analytics, so that their model and tool calls can be traced. We evaluate each agent on its own benchmark, scored against gold answers where they exist and by a model judge for schema mapping and chart quality. Query generation on a public benchmark and routing work well with a current open model, and the data lifecycle management agent solves most of its scenarios. The evaluation exposed defects that only appear when prompts, parsers and serving stacks are tested together, and we report them together with their fixes.
\end{abstract}


\maketitle

\section{Introduction}
\begin{figure}[!htp]
    \centering
    \includegraphics[width=0.78\columnwidth]{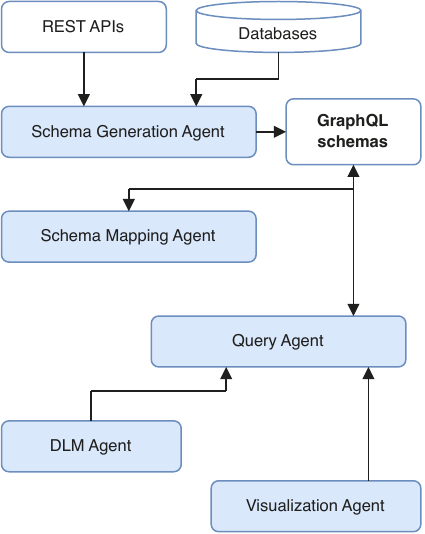}
    \caption{Example workflow in ITOps Observability}
    \Description{Block diagram showing the data flow between the UI chat agent,
               the coordination layer, and the tool-integration layer.}
    \label{fig:intro}
\end{figure}

Large language models (LLMs) and vision language models (VLMs) are now used to automate workflows in data science and analytics. Yet even the most capable models struggle to perform real-world tasks that closely mimic human actions in these domains. These tasks often require operating several professional tools, and benchmarks like Spider2V \cite{cao2024spider2v} show the current limitations of multi-modal agents in automating complete, end-to-end data workflows.

ITOps observability is one such real-world task. By observability we mean the ability to monitor, analyze and act on telemetry, the metrics, logs and traces that systems emit, in order to detect, resolve and prevent issues. The more an organization depends on its IT systems, the more it depends on observability and on automating it. The data for one operator question is often spread across several products, each with its own API, schema and naming conventions.

To automate this, a system must turn an operator's request into queries over each product's schema, and turn the results into charts. Consider the workflow in Figure~\ref{fig:intro}. GraphQL is a query language for APIs in which the client names exactly the fields it wants, so one query can gather related data that would otherwise need several API calls. GraphQL query generation is central to this pipeline, since it translates natural language requests into precise, executable data retrieval operations. However, generating valid and semantically accurate GraphQL queries requires a deep understanding of the underlying schema, which is often heterogeneous and evolving across tools. Generating GraphQL schemas for new sources and mapping between existing ones are therefore prerequisites for data access across products. Data Lifecycle Management (DLM) governs the creation, retention and deletion of data across observability environments. Without it, observability applications may keep unneeded traces, which raises storage costs. Query generation, schema operations and DLM, together with visualization of the results, are the four building blocks of the workflow that we target.

Efforts such as MappedGPT \cite{mappedgpt2023} have demonstrated the feasibility of natural language to GraphQL translation within developer tools, mostly for single queries. For schema mapping, tools like the \texttt{@SchemaMapping} annotation simplify mappings between GraphQL types and resolvers, but require manual setup for each source \cite{springgraphql2023}. In visualization, agents such as nvAgent \cite{wu2024nvagent} and ParaView-MCP \cite{liu2025paraview} show promising results in converting natural language to visualizations or executing complex visual tasks via closed-loop agent workflows, each within a single dataset or visualization environment. Agent frameworks such as AutoGPT \cite{autogpt2023} and LAMBDA \cite{sun2024lambda} outline task decomposition and coordination among agents \cite{solutionsreview2024}, without a focus on observability or GraphQL. Each of these systems covers one step, whether query generation, schema mapping, visualization or agent coordination, and none connects schemas to visualizations for observability data. Work on GraphQL query generation with LLMs shows that models often generate invalid queries when their grounding in the schema is weak \cite{kesarwani-etal-2024-graphql}.

In this work, we address these limitations with ITOps Observability Studio, a multi-agent system that integrates LLMs and VLMs for data science and engineering workflows. It is multi-modal in that its agents work with text and images, and with metrics and traces as semi-structured data such as JSON rather than flat tables. We focus on the Text to Dashboard workflow for ITOps observability, in which an engineer describes in words the dashboard they need. In our system, autonomous agents interpret these requests and retrieve the relevant data from APIs and databases through a GraphQL data access layer. They then design and evaluate dashboards for reliability engineers.
In the following sections, we review related work (Section~\ref{sec:related}), detail our three-tier architecture, inter-agent communication, tool integration and agent analytics (Section~\ref{sec:architecture}), and evaluate the orchestrator's routing and each agent (Section~\ref{sec:studio}). Section~\ref{sec:discussion} discusses the engineering lessons and limitations.

\section{Related Work}
\label{sec:related}

\subsection*{Agentic AI frameworks}

Frameworks such as LangChain and LangGraph enable composition of multi-step LLM workflows using chains, memory, and tool interfaces~\cite{phyniks2025agentic}. LangGraph extends this with graph-based control flows that support loops and branching, better suiting dynamic data workflows. Both rely on pre-indexed or manually curated schemas. CrewAI is a lightweight orchestration layer designed to simulate multi-agent teams with defined roles, goals, and tools~\cite{crewai2024}. AutoGPT pioneered autonomous workflows through iterative task decomposition and plugin integration~\cite{autogpt2023}. Despite its flexibility, it often exhibits hallucinations and inefficiencies due to poor grounding in structured schemas. Microsoft's AutoGen improves multi-agent orchestration via role-based agents and conversational memory~\cite{li2023autogen}. None of these frameworks maps schemas, integrates GraphQL-based APIs or manages the data lifecycle on its own.

Hierarchical frameworks such as HALO (Hierarchical Autonomous Logic-Oriented Orchestration)~\cite{shen2024halo} and Flow~\cite{li2024flow} use planning agents to spawn sub-agents, leveraging search-based strategies for workflow adaptation. Tools like AutoAgent \cite{tang2025autoagent} and AutoAgents \cite{chen2023autoagents} aim to translate natural language instructions into executable agents with no programming required. Many of these agents reason with chain-of-thought prompting \cite{wei2022chain} or interleave reasoning with actions as in ReAct \cite{yaoreact}, which our agents also use. To evaluate such systems, Sapkota et al.~\cite{sapkota2024taxonomy} introduce a taxonomy differentiating narrow task bots from orchestration-based agentic AI, identifying bottlenecks in coordination, hallucination control, and environment alignment. Gabriel et al.~\cite{gabriel2024eval} propose metrics such as Tool-F1 and Structural Similarity Index for assessing performance in tool-driven agent tasks. AIOpsLab \cite{chen2025aiopslab} and ITBench \cite{jha2025itbench} provide environments and tasks for evaluating AI agents on cloud operations, site reliability and FinOps tasks. RCAgent \cite{wang2024rcagent} applies tool-augmented LLM agents to root cause analysis in the cloud. NL2KQL \cite{tang2024nl2kql} translates natural language into Kusto queries over log and telemetry tables, a task close to our GraphQL query generation. These works benchmark agents on incident and operations tasks, while ours address data access, schema integration and visualization across several observability products.

Two open protocols cover most agent communication. The Model Context Protocol (MCP), introduced by Anthropic in November 2024~\cite{anthropic2024mcp}, connects LLM applications to tools and data over JSON-RPC. Its 2026-07-28 revision~\cite{mcp2026spec} makes the core stateless and moves further functionality into opt-in extensions, among them \emph{Tasks} for long-running operations. The Agent2Agent (A2A) protocol~\cite{google2025a2a,a2a2026spec}, introduced by Google in April 2025, standardizes how one agent discovers another through an agent card and delegates tasks to it, and has reached version~1.0. IBM's Agent Communication Protocol (ACP)~\cite{ibm2025acp}, a REST-based protocol, merged into A2A under the Linux Foundation in 2025. Our system uses MCP between agents and tools and A2A between agents, which matches this division of labor. Agent Skills~\cite{agentskills2025} package instructions, scripts and resources that an agent loads only when needed, and complement MCP tools with procedural knowledge.

\subsection*{API Function Calling and Sequencing}
Several recent works improve the use of APIs in agent workflows. Saha et al.~\cite{saha2024sequential} focus on GraphQL schema design for sequential API function calling, enabling agents to chain requests in a structured manner. Mandal et al.~\cite{mandal2024graph} propose graph-based approaches for sequencing APIs using an API dependency graph, which helps agents reason over available endpoints. Basu et al.~\cite{basu2024nestful} study nested sequences of API calls, which arise in IT automation when several agents interact. Abdelaziz et al.~\cite{abdelaziz2024granitefunctioncalling} introduce Granite Function Calling to standardize function invocation across heterogeneous APIs. Basu et al.~\cite{basu-etal-2024-api} also present a benchmark corpus for training language models on API detection and sequencing. Finally, Chen et al.~\cite{chen2024apigen} propose APIGen, a method to help agents to automatically discover and leverage suitable APIs for a given task.

\subsection*{Tool Use and API Integration}
AnyTool \cite{anytool} presents self-reflective, hierarchical agents for large-scale API calls. ToolLLM \cite{toolllm} shows how LLMs can use many real-world APIs, and Gorilla \cite{patil2023gorilla} trains LLMs with API information. API-Bank \cite{li2023api} and SEAL \cite{kim2024sealsuiteevaluatingapiuse} evaluate how LLMs use APIs, and Metatool \cite{huang2023metatool} addresses whether and which tools to use. RestGPT \cite{song2023restgpt} connects LLMs with RESTful APIs, as our data access layer does through GraphQL. TaskMatrix.AI \cite{liang2024taskmatrix} completes tasks by connecting foundation models with millions of APIs.

\subsection*{API Mapping and Documentation}
Several works provide foundational techniques relevant to schema mapping agents. Lu et al.~\cite{lu2017learning} learn API mappings from documentation, and Liu et al.~\cite{liu2023recommending} recommend APIs with a knowledge graph that relates semantically similar endpoints. Shao et al.~\cite{shao2022cross} map APIs across platforms, and Moon et al.~\cite{moon2024api} recommend API-to-API specifications as APIs evolve. Wittern et al.~\cite{wittern2018graphqlwrappers} generate GraphQL wrappers for REST-like APIs from their OpenAPI descriptions, the same input our schema generation agent uses, with fixed rules rather than an LLM. Our schema mapping agent pursues the same goal for GraphQL schemas rather than APIs (Section~\ref{sec:schema-mapping}).

\subsection*{LLM Fine Tuning}
Our agents use prompted models without fine-tuning, and Section~\ref{sec:graphql-query} names fine-tuning as future work. Full fine-tuning trains every parameter of a model and needs substantial compute, memory and training time. Instruction fine-tuning~\cite{zhang2024instructiontuninglargelanguage} trains a model on labeled instruction and response pairs so that it follows complex instructions more closely. Parameter-efficient methods train far fewer parameters. LoRA~\cite{hu2021loralowrankadaptationlarge} adds small adapters to a frozen model, and QLoRA~\cite{dettmers2023qloraefficientfinetuningquantized} also quantizes the model to 4 bits, so that models with more than 65B parameters can be tuned on a single GPU.

\subsection*{Visualization}
\label{sec:rw-visualization}

Natural Language Interfaces for Data Visualization (V-NLIs) let users create charts by describing them in natural language (NL). Existing approaches can be broadly categorized into (1) NL-guided systems, where visualizations are shaped by user queries, and (2) data-driven systems, which generate visualizations directly from data without NL input. Early NL-guided methods relied on rule-based NLP pipelines that mapped parsed queries to visualization parameters using grammar rules and similarity metrics such as Levenshtein and Wu-Palmer distances. For instance, BOLT~\cite{Srinivasan2023BOLT} used rigid intent templates but struggled with vague or ambiguous prompts. Data-driven systems like Calliope~\cite{9222368} and Dashbot~\cite{9906971} applied Monte Carlo Tree Search and reinforcement learning, respectively, to generate dashboards from data alone. These approaches lacked semantic understanding, could not support hierarchical data, and produced static visualizations that were hard to interpret.

Recent advances use LLMs to address these limitations. LIDA~\cite{dibia2023lidatoolautomaticgeneration} employs a modular LLM-assisted pipeline for summarization, goal inference, and visualization generation, achieving a 3.5\% Visualization Error Rate (VER) and strong GPT-4-based quality scores. NL4DV-LLM~\cite{sah2024generatinganalyticspecificationsdata} handles ambiguous queries by inferring intents from prompt-engineered queries, reaching 87.02\% accuracy on 740 annotated examples, albeit with higher latency. Chat2Vis~\cite{10121440} uses LLMs to generate Vega specifications from prompt examples, but suffers from high output variance and limited stylistic control.

Despite these improvements, most methods lack robust evaluation. Many rely on case studies or subjective metrics, limiting reproducibility. VisEval~\cite{chen2024viseval} addresses this by introducing an automated evaluation framework that quantifies visualization quality along three axes, namely Validity (checks for crashes), Legality (verifies alignment with query intent), and Readability (assessed via GPT-4V). It provides a reproducible benchmark, revealing common pitfalls like invalid code, incorrect encodings, and poor layout quality. Our evaluation of the visualization agent follows VisEval~\cite{chen2024viseval} and adds a usability criterion for SRE decision-making.

\section{Three-Tier Reference Architecture}
\label{sec:architecture}

We build the \textit{Observability Studio} as a multi-agent system, so that each task has its own specialized agent. Multi-agent systems have also been shown to simulate human strategic behavior more accurately than a single LLM agent \cite{sreedhar2024simulatinghumanstrategicbehavior}. Building on specialized agents, we introduce a three-tier reference architecture that coordinates multiple agents using an agent-to-agent communication protocol. While our implementation uses the Agent2Agent (A2A) protocol \cite{google2025a2a}, the design adapts to any protocol with structured inter-agent messaging. The architecture coordinates data workflows across heterogeneous IT systems. Each agent, tool and protocol can be replaced on its own, and agents that change data, such as the DLM agent, ask the user before they apply a sampling policy.

\begin{figure*}[t]
    \centering
    \includegraphics[width=0.9\textwidth]{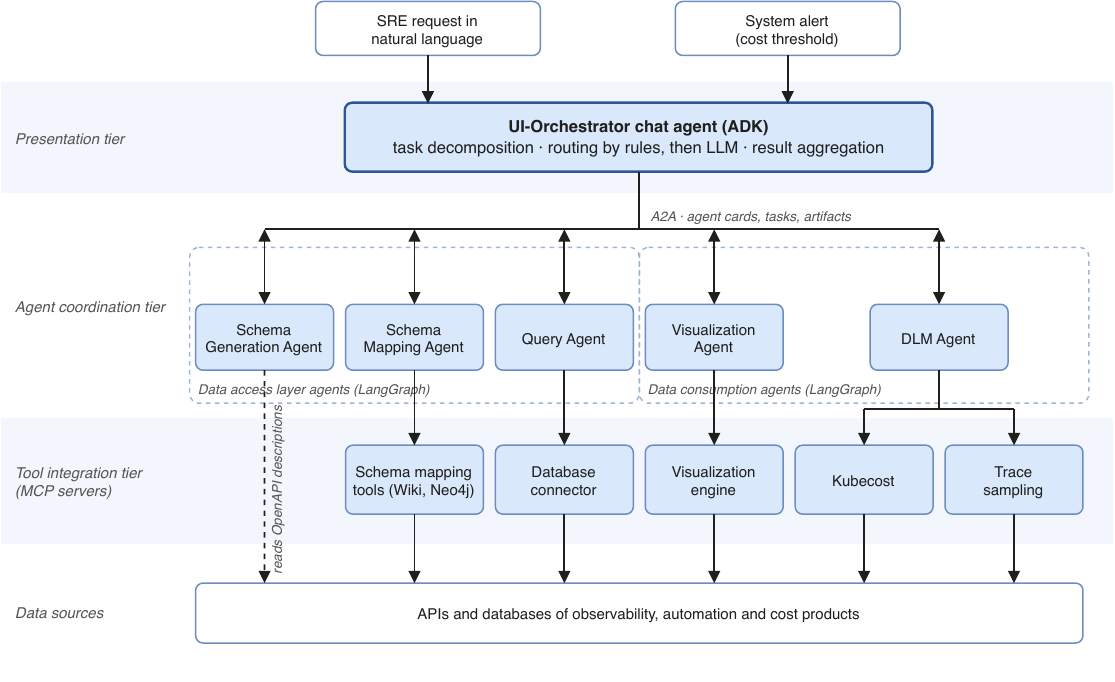}
    \caption{The three-tier reference architecture of ITOps Observability Studio, with the tools the agents call and the data sources behind them}
    \Description{Layered diagram. Requests from an SRE and system alerts reach the UI-Orchestrator chat agent in the presentation tier, which talks over A2A to five LangGraph agents in the agent coordination tier, grouped into data access and data consumption agents. Below them, MCP servers in the tool integration tier connect to the APIs and databases of observability, automation and cost products.}
    \label{fig:architecture}
\end{figure*}

\subsection*{Architectural Overview}

The architecture has three tiers. The presentation tier is an ADK-based UI chat agent that mediates interactions between the user and the backend agents. The agent coordination tier holds specialized LangGraph agents that collaborate via a pluggable, protocol-agnostic communication layer. The tool integration tier consists of MCP servers that offer domain-specific functionalities through external tools and APIs. Figure~\ref{fig:architecture} shows the three tiers, the agents' tools and the data sources behind them. The tiers meet only through an abstract communication layer, so an agent can move to another agent framework or protocol without changes to other agents.

\subsection*{Agent Ecosystem}

We implement six specialized agents, grouped by function into data access agents, data consumption agents and an orchestrator. Three agents form the data access layer. The GraphQL Query Agent extends the sequential API function calling approaches \cite{saha2024sequential,mandal2024graph} by generating contextually-aware GraphQL queries from natural language descriptions. Unlike traditional API sequencing methods, this agent uses LangGraph's reasoning to understand the query requirements and to generate GraphQL operations over the target GraphQL schema. The GraphQL Schema Generation Agent generates GraphQL schemas from the OpenAPI descriptions of data sources, so new sources can join the federated GraphQL layer, the single endpoint through which all sources are queried, as their APIs evolve. The GraphQL Schema Mapping Agent clusters schemas by domain and structure before it maps them (Section~\ref{sec:schema-mapping}), extending embedding-based approaches \cite{hättasch2022its} and foundation model methods \cite{narayan2022foundation}. Domain classification and structural analysis give it more context-aware mappings between heterogeneous schemas.

Two agents consume this data. The Visualization Agent builds on the visualization frameworks of Section~\ref{sec:rw-visualization}, combining data-driven generation with natural language guidance. Unlike LIDA \cite{dibia2023lidatoolautomaticgeneration} and Chat2Vis \cite{10121440}, this agent works on the GraphQL data layer, so data access and transformation happen within the visualization pipeline. The Data Lifecycle Management (DLM) Agent recommends and applies trace pruning policies (Section~\ref{sec:dlm}). It coordinates with data access agents to understand data lineage and with the visualization agent to present analyses of cost impact to the user.

The ADK-based UI-Orchestrator Chat Agent is the orchestrator and user interface, using the Agent Development Kit (ADK) to provide a conversational interface while coordinating multi-agent workflows. It decomposes tasks, routes them and aggregates the results. Routing is decided in two stages. It first applies keyword rules, and if no rule matches, its LLM chooses an agent from the agent cards. We evaluate both routing stages in Section~\ref{sec:routing}.

\subsection*{Agent-to-Agent Communication}

In this work, we primarily use the A2A protocol \cite{google2025a2a} for inter-agent communication. Each agent publishes an agent card, a short machine-readable description of its skills and how to reach it, and the orchestrator sends tasks to agents and receives messages and artifacts back. The protocol keeps agents loosely coupled, and our architecture uses it to support four coordination patterns. In sequential coordination, the UI-Orchestrator agent routes user requests to the appropriate specialized agents in sequence. In parallel processing, multiple agents can process different aspects of a request simultaneously, for example, schema generation and mapping can occur in parallel while the visualization agent prepares visualization templates. In hierarchical delegation, the agents that the orchestrator calls may delegate subtasks to other agents. In collaborative refinement, agents can iteratively refine results through multiple interaction rounds, for example, the schema mapping agent may request additional context from the query agent to improve mapping accuracy. Algorithm~\ref{alg:cost-analysis-workflow} shows a multi-agent cost analysis workflow.

\subsection*{Tool Integration}

The third tier implements tool integration through MCP servers \cite{anthropic2024mcp}, providing standardized access to external capabilities. This design makes tools reusable across multiple agents while maintaining clear separation between agent logic and tool implementation. The Kubecost MCP server provides cost analysis for Kubernetes clusters, using Kubecost, a cost monitoring tool. The trace sampling MCP server implements various trace sampling algorithms and evaluation metrics. The database connector MCP server enables multi-database query execution and schema introspection. The visualization engine MCP server processes Vega-Lite specifications and generates multiple output formats. The tools reach observability, automation and cost products such as Instana, Turbonomic, Concert and Kubecost, each with its own API and schema.

\begin{algorithm}[H]
\caption{Multi-Agent Cost Analysis and Optimization}
\label{alg:cost-analysis-workflow}
\begin{algorithmic}[1]
\REQUIRE User request for cost analysis with visualization and optimization
\ENSURE Response with cost data, visualizations, and optimization recommendations

\STATE TaskDecomposition $\leftarrow$ UI-Orchestrator parses request and identifies the required capabilities, namely cost analysis, data visualization and trace optimization

\STATE SchemaDiscovery $\leftarrow$ Schema Generation Agent discovers relevant Kubecost and observability schemas

\STATE SchemaIntegration $\leftarrow$ Schema Mapping Agent creates unified schema mappings between cost and trace data sources

\STATE QueryGeneration $\leftarrow$ GraphQL Query Agent generates queries for cost and trace data retrieval

\STATE VisualizationCreation $\leftarrow$ Visualization Agent processes the data and generates visualizations

\STATE CostOptimization $\leftarrow$ DLM Agent analyzes cost patterns and provides trace sampling recommendations

\STATE ResultAggregation $\leftarrow$ UI-Orchestrator Agent combines all artifacts into one response

\RETURN Integrated response containing cost analysis, visualizations, and optimization recommendations
\end{algorithmic}
\end{algorithm}

In Algorithm~\ref{alg:cost-analysis-workflow}, the orchestrator first splits the user's cost question into cost analysis, data visualization and trace optimization. The three data access agents then work in sequence. They discover the Kubecost and observability schemas, map cost data to trace data, and generate the queries that retrieve both. The visualization agent and the DLM agent consume the retrieved data, the first as charts and the second as trace sampling recommendations, and the orchestrator combines their artifacts into one response. Each of these steps is an A2A task sent to one agent, and each agent reaches its data through the MCP servers of the tool integration tier.

\subsection*{Agent Analytics}
\label{sec:analytics}

The agents are themselves software that needs observing. Five agents, the orchestrator and the query, schema mapping, visualization and DLM agents, initialize an agent-analytics SDK at start-up. The SDK builds on OpenTelemetry and records each agent's LLM calls and tool calls as traces, which it writes to a per-agent log.

These traces let us attribute latency and model calls to individual agents, and connect a wrong answer to the routing decision, tool call or model call that produced it. Each agent writes its own log, so following a request across agents means joining the logs of the agents it reached, and propagating the trace context over A2A would make this direct. The same signals map onto the OpenTelemetry GenAI semantic conventions \cite{otel2025genai}, a shared vocabulary for the telemetry of AI applications. Under these conventions, an agent invocation, a model call and a tool call each become a span (one timed step of a trace), and token usage and operation duration become metrics. The agents thus emit the same telemetry as the systems they observe, and the same tools can analyze them.

\section{ITOps Observability Studio}
\label{sec:studio}

In this section, we present a case study of how the system described in Section~\ref{sec:architecture} works in the ITOps observability domain. We evaluate each agent separately, on its own task and metric, with gold answers wherever they exist. Table~\ref{tab:summary} summarizes the results, and the subsections below detail each agent. Its routing row is the LLM router, and the deployed hybrid router scores slightly lower on both of these measures (Figure~\ref{fig:routing}). The original implementation used Granite-3.x models, no longer served, so we rerun every agent with current models on IBM watsonx. The main model is Llama-3.3-70B-Instruct, and the smaller IBM model compared is Granite-4-H-Small. Decoding is greedy. Each evaluation drives the deployed agent code in-process, and each agent's section notes any change that running it needed, such as a repaired parser.

\begin{table*}[t]
\centering
\setlength{\tabcolsep}{4pt}
\begin{tabular}{llllll}
\toprule
\textbf{Agent} & \textbf{Benchmark} & \textbf{$n$} & \textbf{Model} & \textbf{Metric} & \textbf{Result} \\
\midrule
Orchestrator (routing) & ITOps requests & 351 & Llama-3.3-70B & accuracy / macro-F1 (LLM) & 0.943 / 0.827 \\
GraphQL query & NL-to-GraphQL test & 1,164 & Llama-3.3-70B & schema-valid / exact (zero-shot) & 85.7\% / 58.4\% \\
Schema generation & OpenAPI $\rightarrow$ GraphQL & 3 products & Llama-3.3-70B & field F1 / type F1 & 0.67 / 0.27 \\
Schema mapping & product schema pairs & 2,439 mappings & LLM judge & joinable / equivalent & 45.51\% / 24.56\% \\
Visualization & SRE queries & 30 & Llama-3.3-70B & VER / legality (gen., spec.) & 0\% / 97.8\%, 62.3\% \\
DLM (trace pruning) & constructed scenarios & 30 & Llama-3.3-70B & task accuracy & 0.733 \\
\bottomrule
\end{tabular}
\caption{Summary of the per-agent evaluation}
\label{tab:summary}
\end{table*}

\subsection*{ITOps Observability Dataset}

The ITOps Observability dataset contains 351 natural-language requests of the kind a site reliability engineer (SRE) sends to the orchestrator, stored as JSONL. Each request carries a task type, which names the agent that should handle it. Of the 351 requests, 289 are schema mapping requests, 30 GraphQL query requests, 29 visualization requests and 3 trace pruning requests. The task types were assigned by the generator that produced the dataset rather than by human annotators. Existing resources typically address only a single stage of this workflow. \cite{kesarwani-etal-2024-graphql} focuses on natural-language to GraphQL translation and \cite{saha2024sequential} on sequential function calling with GraphQL, whereas our requests span several cooperating agents.

\subsection*{Routing}
\label{sec:routing}

We evaluate how the orchestrator routes the 351 requests to the four specialized agents. The \emph{rule} router runs the orchestrator's keyword scoring function unchanged. The \emph{LLM} router replicates the ADK path, with the orchestrator's instruction and agent cards as system prompt, and predicts the agent named in its first \texttt{send\_message} call. The \emph{hybrid} router is the orchestrator's behavior when all agents are reachable, rules first, then the LLM if no rule matches.

\begin{figure}[t]
\centering
\includegraphics[width=\columnwidth]{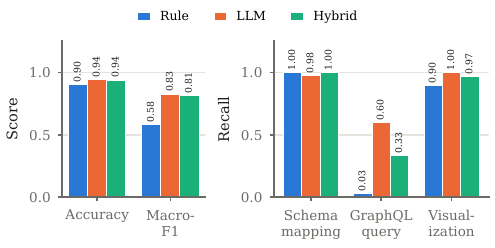}
\caption{Routing of the ITOps requests to the specialized agents, overall and by agent}
\Description{Bar charts of accuracy and macro-F1 for the rule, LLM and hybrid routers, and their recall on schema mapping, GraphQL query and visualization requests.}
\label{fig:routing}
\end{figure}

Figure~\ref{fig:routing} shows the results, with each router's recall on the three larger classes. Because the classes are unbalanced, macro-F1, which weights every agent equally, is the more informative measure. Accuracy is high for every router because almost all of the 289 schema mapping requests start with ``map'', which the keyword rule catches. The weak class is GraphQL query. Most of them are phrased as display requests (``Show me \ldots'', ``Display distribution \ldots''), which the rules send to the visualization agent or do not route at all. The LLM recovers most of them, but in the hybrid router the rules override it wherever they fire, lowering GraphQL recall. Trace pruning has only three requests, too few for a per-class number.

\subsection*{GraphQL Query Agent}
\label{sec:graphql-query}

The current GraphQL Query Agent is built using the \texttt{LangGraph} framework. The agent uses a ReAct-style planning loop to translate natural language queries into structured GraphQL calls. Each action is iteratively selected based on intermediate reasoning traces. The agent uses schema-aware tool definitions for GraphQL operations, which are loaded and parsed to tell the agent about the types, fields and relations, including nested and paginated structures.

\subsubsection*{Illustrative Example}
Consider the following request.
\begin{quote}
\emph{``Give me the area in square kilometers and the hosts of the farm competition for city with ID 1.''}
\end{quote}

\noindent{The relevant part of the schema is as follows.}
\begin{quote}
    \texttt{City} implements interface \texttt{census\_details}
    \texttt{City} has field \texttt{area\_km\_2} and a relationship with
    \texttt{farm\_competition}
    \texttt{Farm\_competition} contains the \texttt{hosts} field
\end{quote}

\noindent{The agent generates the following GraphQL query.}

\begin{verbatim}
query {
  city(city_id: 1) {
    area_km_2
    farm_competition {
      hosts
    }
  }
}
\end{verbatim}

The agent reasons over nested schema relationships, finding the relevant fields across types and combining them into one query. It follows the relationship from \texttt{City} to \texttt{farm\_competition} instead of a second query. It also selects only the two requested fields.

\subsubsection*{Evaluation}

We use the test split of our NL-to-GraphQL dataset, built from WikiSQL and CoSQL schemas, with 1,164 questions over 13 schemas and 10,940 training pairs. We compare zero-shot generation with 3-shot prompting, with examples retrieved from the training split by TF-IDF similarity. We report whether the query parses, validates against its schema, and matches the gold query exactly after normalization. The dataset ships schemas and queries but no data, so we cannot measure execution accuracy. Its schemas are general-domain rather than observability schemas, so the results measure query generation in general and not on ITOps products.

\begin{figure}[t]
\centering
\includegraphics[width=\columnwidth]{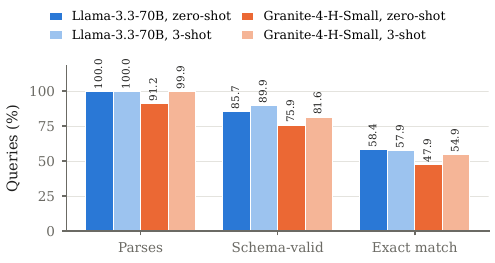}
\caption{Natural language to GraphQL generation on the test split}
\Description{Bar chart of the share of generated queries that parse, validate against the schema and match the gold query, for two models with and without examples.}
\label{fig:graphql-query}
\end{figure}

Figure~\ref{fig:graphql-query} shows the results. With Llama-3.3-70B, 86--90\% of queries are schema-valid and 58\% match the gold query exactly. Examples help validity for both models, and exact match only for Granite. Exact match understates correctness, since a query can be equivalent to the gold query without being identical.

Prompted models are not always reliable on domain-specific schemas. Parameter-efficient fine-tuning such as LoRA~\cite{hu2021loralowrankadaptationlarge} and QLoRA~\cite{dettmers2023qloraefficientfinetuningquantized} could adapt a model to our schemas on modest hardware, and a fine-tuned model could be shared by several agents that work over structured schemas. We leave this to future work.

\subsection*{GraphQL Schema Generation Agent}
\label{sec:schema-generation}

The schema generation agent turns an OpenAPI specification into GraphQL schema definitions. It splits the specification into chunks of at most four operations, prompts the LLM to write their types and query fields, and assembles the result. We compare the generated schemas with the hand-written GraphQL schemas maintained for three products, the only ones with both an OpenAPI specification and such a schema. The generator sees exactly the operations that the hand-written schema covers. We measure whether each generated file parses and builds, and match the object types and the fields within matched types against the hand-written schema.

\begin{table}[t]
\centering
\begin{tabular}{lcc}
\toprule
\textbf{Measure} & \textbf{Current agent} & \textbf{Earlier output} \\
\midrule
Parses & 0.87 & 0.42 \\
Builds (strict / lenient) & 0.28 / 0.87 & 0.00 / 0.19 \\
Type P / R / F1 & 0.29 / 0.26 / 0.27 & 0.01 / 0.04 / 0.02 \\
Field P / R / F1 & 0.63 / 0.72 / 0.67 & 0.20 / 0.17 / 0.18 \\
\bottomrule
\end{tabular}
\caption{Generated schemas against the hand-written schemas}
\label{tab:schema-generation}
\end{table}

Table~\ref{tab:schema-generation} shows the macro mean over the three products. The earlier output is the stored output of the first version of the agent, and the lenient build declares the \texttt{JSON} scalar before building. Every strict build failure comes from this undeclared scalar, which the prompt itself asks the model to use. Within matched types, fields agree well, but types agree poorly. The hand-written schemas are a curated query layer that renames types, adds connections and keeps only part of each API. These scores therefore measure agreement with a human design rather than with correctness.

\subsection*{GraphQL Schema Mapping Agent}
\label{sec:schema-mapping}

Schema mapping involves defining correspondences between the schemas of different APIs to create a unified, queryable interface. It aligns the fields of a source schema to those of a target schema, so that data types, formats and meanings are compatible. When integrating multiple APIs, schema mapping ensures that data from diverse sources, possibly with varying structures, naming conventions or data types, can be accessed through a single federated GraphQL endpoint. Approaches range from algorithmic rule-based methods to machine learning and language models.

DynaMap \cite{mazilu2022schema} uses dynamic programming to search possible mappings, while \cite{hättasch2022its} applies embedding-based similarity for table and attribute matching. Valentine \cite{koutras2021valentine} benchmarks them against gold matches. SMUTF \cite{zhang2025smutf} combines rule-based signals and semantic embeddings in an XGBoost classifier. The Learned Schema Mapper \cite{zhang2023schema} fine-tunes a BERT encoder on schema metadata. Magneto \cite{liu2024magneto} generates candidates with small language models and reranks them with an LLM, and ISResMat \cite{du2024in} fine-tunes a language model on pairwise samples. \cite{narayan2022foundation} phrases schema mapping as a prompt to a foundation model. Retrieval-augmented methods add context, either metadata documents from the target schema \cite{sheetrit2024rematch} or knowledge graph facts \cite{ma2025knowledge}, and GraphRAGToolFusion \cite{lumer2025graphragtoolfusion} combines knowledge graphs with vector retrieval for tool selection.

However, these approaches do not explicitly organize or cluster schemas based on API domain, functionality, or structural complexity. AnyTool \cite{anytool} introduces a hierarchical retrieval architecture with three levels of agents. A meta-agent interprets user intent and routes queries, category-agents specialize in specific domains to narrow down relevant tools, and tool-agents assess and execute individual APIs. Inspired by this architecture, we built a modular, hierarchical framework for automating schema mapping. A toolkit of functions implements the mapping pipeline. When a mapping request arrives, the system first analyzes the source and target schemas to determine their domain and structural characteristics, and clusters them by domain and functionality. It then prompts an LLM to predict field correspondences. The responses are filtered and converted into a structured format, justifications are added, and mappings are ranked by semantic similarity to the source fields.

For example, when given a natural language query such as ``Map concert vulnerabilities to turbo entities'', the agent first computes semantic similarity scores between the query and all available schemas to shortlist the top ten candidates. From these, the source and target schemas are selected by an LLM based on the user's intent. The toolkit is then executed on the selected schemas to identify relevant types and fields, and the resulting mappings are returned. Finally, the agent can invoke additional tools such as a Neo4j tool or a Wiki tool. The Neo4j tool stores the mappings in a graph database by hierarchically organizing them as product $\rightarrow$ type $\rightarrow$ field, with field-level mappings represented as edges. Alternatively, the Wiki tool generates hierarchical documentation pages for each schema and embeds hyperlinks to represent the discovered mappings. Two example mappings and their justifications follow.

\smallskip
\noindent\path{VulnerabilityAssociationComponent.issues.id} $\rightarrow$\\
\path{turbo_Action.actionID}\\
Both fields represent unique identifiers, where \texttt{id} identifies an issue and \texttt{actionID} identifies an action.

\smallskip
\noindent\path{VulnerabilityAssociationComponent.issues.fixed_on} $\rightarrow$\\
\path{turbo_Action.clearTime}\\
These fields relate to resolution timestamps, where \texttt{fixed\_on} indicates when an issue was resolved and \texttt{clearTime} when an action was cleared.
\smallskip

The Schema Mapping Agent was tested on schema pairs from heterogeneous systems and evaluated by an LLM-as-a-judge, Mistral Large. The judge scores each mapped pair between $0$ and $1$ for joinability and for equivalence. Joinability indicates whether the output of one field can be used to populate or enrich the other, while equivalence indicates whether the two fields represent the same core function. A pair with a score above $0.5$ counts as correct, and accuracy is the share of correct mappings. The test set holds $289$ schema mapping requests, whose schema pairs were selected by hand from the API descriptions. For these requests the agent produced $2{,}439$ field mappings, all passed to the judge (Table~\ref{tab:summary}). The judge finds fewer than half of the mappings joinable and about a quarter are equivalent, with mean scores of 0.43 and 0.34. A judge has no gold answers, so the numbers estimate rather than measure correctness, and a gold-scored evaluation is future work.

\subsection*{Data Lifecycle Management (DLM) Agent}
\label{sec:dlm}

The DLM agent builds on work in which LLMs call external tools. Toolformer \cite{schick2023toolformer} demonstrated how LLMs can teach themselves to call APIs. AnyTool \cite{anytool} proposed a hierarchical, self-reflective agent that uses GPT-4's function calling to orchestrate access to over $16{,}000$ APIs. MARCO \cite{shrimal2024marco} introduced a framework for multi-agent real-time orchestration, with guardrails and symbolic planning, for the reliable execution of tasks across business workflows.

The Data Lifecycle Management (DLM) agent is designed for efficient trace monitoring and pruning. It integrates with monitoring systems to track costs across clusters, namespaces, applications, and services. Using trace volume management tools, the agent identifies policies for trace reduction, recommends these policies to the user, and executes pruning actions accordingly. The agent is built with LangGraph, with tools deployed as MCP servers.

\begin{figure}[!t]
  \centering
  \includegraphics[width=\linewidth]{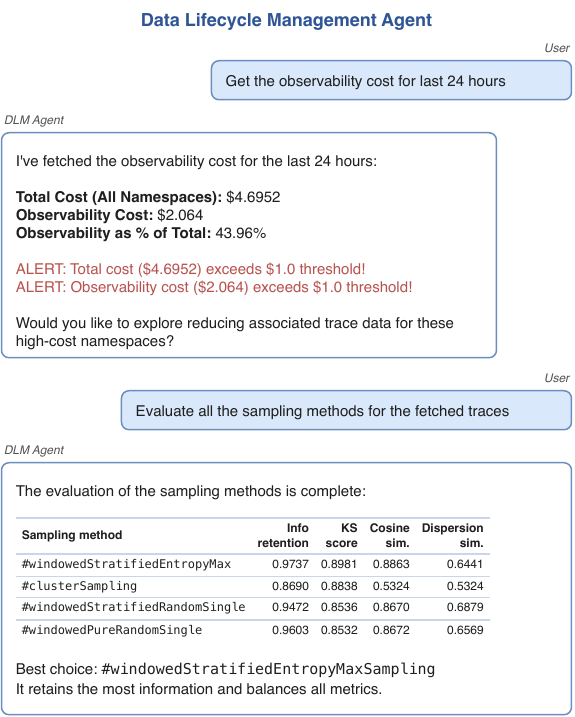}
  \caption{A sample interaction with the DLM agent}
  \Description{Chat transcript. The user asks for the observability cost of the last 24 hours, and the agent reports total and observability cost with two threshold alerts. The user asks to evaluate the sampling methods, and the agent returns a table of four methods and recommends windowed stratified entropy-max sampling.}
  \label{fig:dlm-architecture}
\end{figure}

These data points form the input context for the DLM Agent, which connects to the other tiers through A2A. The agent reaches the data access layer and the DLM tool through two MCP servers. A Kubecost server provides allocation and cloud cost data and cost alerts. A trace sampling server evaluates and applies sampling methods, which keep a representative subset of traces and drop the rest, cutting storage cost at the price of some detail. The DLM Agent can be triggered in two ways, system-initiated, where Kubecost generates alerts when telemetry volume causes high cloud costs, and user-initiated, where a user asks to reduce costs.

Based on the telemetry data and cost, the agent evaluates trade-offs between cost and information retention. This is facilitated by the Trace API, which internally uses a \emph{Trace Volume Manager} (TVM). The TVM supports multiple trace sampling algorithms and is designed to be extensible, allowing additional sampling strategies to be integrated as needed. The agent selects the recommendation that balances cost savings with an acceptable loss of trace visibility, deploys the policy and communicates with the tool for trace pruning. Figure~\ref{fig:dlm-architecture} shows a sample interaction with the agent. Further tools or logic modules can be plugged in through MCP servers.

\subsubsection*{Evaluation}

We constructed 30 scenarios, six for each query type, of which seven are multi-turn. The agent code and both MCP servers run unchanged, with only their HTTP calls to Kubecost and the TVM replaced by deterministic mocks. The mocks make the best sampling method depend on the requested rate. A scenario counts as correct only if the sequence of tool calls, the arguments (window, sampling rate, method) and the key fact in the answer are all correct.

\begin{figure}[t]
\centering
\includegraphics[width=\columnwidth]{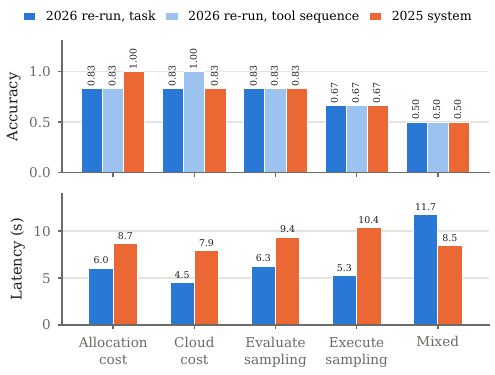}
\caption{DLM agent accuracy and latency by query type, in the re-run and in the original system}
\Description{Bar charts of task accuracy, tool-sequence accuracy and latency for five query types, comparing the re-run with the original system.}
\label{fig:dlm}
\end{figure}

Figure~\ref{fig:dlm} shows the results, with latency per scenario summed over turns. The agent solves 22 of 30 scenarios and calls the expected tools in 23. When it executes a policy, it chooses the best sampling method in 5 of 9 cases. Most failures come from wrong tool calls, such as the cloud cost tool when allocation data is needed, or only one of two needed tools. Another is a first sampling evaluation with an invalid rate of 0. It also sometimes stops after evaluation without executing, which is consistent with the prompt's rule to ask the user before acting. With six scenarios per type, one scenario changes a type's accuracy by 0.17. The original system, measured in 2025 on its own scenarios and gateway, had similar accuracy per query type and was slower on all but mixed requests.

On watsonx, Llama-3.3's chat template ignored tool results whenever a system message was present, and the model re-issued the same tool call until LangGraph's recursion limit. The original deployment, behind a different gateway, did not show this. We moved the system prompt, verbatim, into the first user turn. The agent's behavior thus depends on the serving stack as well as the model.

\subsection*{Visualization Agent}
\label{sec:visualization}

The Visualization Agent automatically generates and assesses data visualizations in response to natural language queries about IT observability. It has three stages, preprocessing, visualization and postprocessing. Preprocessing cleans the input dataset by removing columns with more than $50\%$ missing values and dropping redundant features. Redundancy is judged by an LLM, which reasons over the schema to find semantically overlapping fields. The visualization stage translates natural language queries into Vega-Lite 5 specifications, a format in which a chart is described declaratively as JSON, using Chain-of-Thought (CoT) prompting. The prompt includes syntactic constraints and a rationale for chart selection, inspired by the ReAct framework. The LLM is provided with the data schema, including column names and data types. The system supports five chart types, namely bar, line, area, pie and scatter plots. Postprocessing validates and evaluates the generated visualizations. A syntactic check ensures that the Vega-Lite specification is compatible with the dataset, and valid specifications are then rendered and passed to a VLM for semantic evaluation.

\subsubsection*{Evaluation}

We assess the agent with one syntactic and five semantic metrics. The Visualization Error Rate (VER) is the proportion of Vega-Lite specifications that fail to render due to syntax or data mismatches. For the semantic evaluation, a VLM assesses the quality of rendered charts across five dimensions. The first four come from prior NL2VIS work~\cite{chen2024viseval}, and the fifth captures domain-specific usability in ITOps. Legality (yes or no) asks whether the chart matches the intended type with a correct logical mapping. Layout (yes or no) asks whether the layout is free of clipping, overlapping or a misplaced legend. Scale and ticks (0--5) asks whether the axes are appropriately scaled and labeled for interpretability. Readability (0--5) assesses visual clarity, including titles, labels, fonts, colors and spacing. Usability (0--5) evaluates whether the visualization supports the real-world decision-making of SRE teams.

The original implementation used \path{ibm/granite-3-2-8b-instruct}~\cite{IBM_Granite_3_2_8B_Instruct_2025} for the preprocessing and visualization stages and \path{pixtral-12b-2409}~\cite{agrawal2024pixtral12b} as the evaluation VLM. Neither is still served. We therefore run the deployed pipeline with Llama-3.3-70B and Granite-4-H-Small as text models and Mistral Medium as the judge, keeping the original prompts. We curated $30$ natural language queries from interviews with SREs, of which $10$ are generic and $20$ are specific. Generic queries are broad requests that often need several charts, such as \textit{``Analyze the overall health and performance of the application.''} Specific queries are targeted questions addressable via a single chart, such as \textit{``Create a scatter plot of mean latency vs. calls per second to identify performance bottlenecks.''} The data are samples of Instana application and endpoint metrics.

The deployed specification extractor requires a fenced \verb|```json| block, but the prompt tells the model not to emit one. Both current models follow the prompt, so as shipped the pipeline would produce no charts. For the evaluation we use the brace-matching extractor in the same module, which recovers every specification. The fix is to align the parser with what the prompt asks for.

\begin{figure*}[t]
\centering
\includegraphics[width=\textwidth]{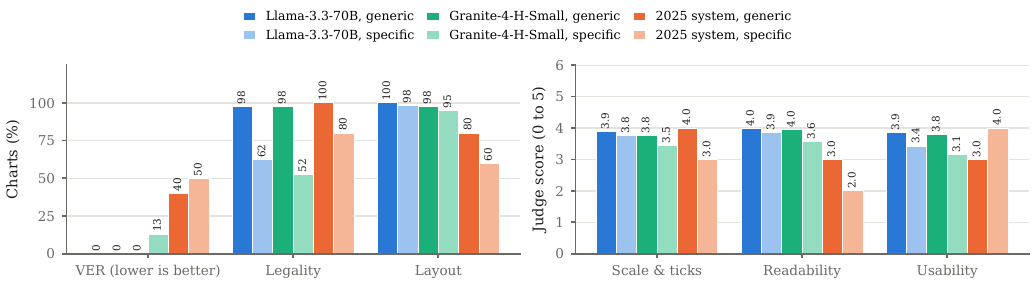}
\caption{Visualization agent performance across query types for both current models and the original system}
\Description{Bar charts of visualization error rate, legality and layout in percent, and judge scores for scale and ticks, readability and usability, for generic and specific queries.}
\label{fig:viz-evaluation}
\end{figure*}

Figure~\ref{fig:viz-evaluation} shows the results. Llama-3.3-70B produced 45 specifications for the generic queries and 61 for the specific ones, and Granite-4-H-Small 47 and 46. The figure also shows the scores of the original system, with Granite-3.2-8B judged by Pixtral-12B on the data of the time, so the two are not directly comparable. VER is zero for every model and query type except Granite-4-H-Small on specific queries, and the small, clean sample data probably explain much of this. Generic queries score higher than specific ones on legality and usability. Many specific queries ask for chart types the prompt forbids, such as histograms, box plots and dual axes, so the model substitutes an allowed type, which the judge marks illegal.

\subsubsection*{Qualitative Results}

\begin{figure}
    \centering
    \includegraphics[width=\linewidth]{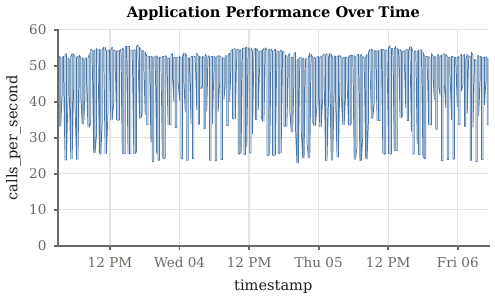}
    \caption{Time-series visualization of application call volumes, indicating normal operational behavior}
    \Description{Line chart of application call volume over time.}
    \label{fig:calls-distribution}
\end{figure}

Upon receiving a generic query, the orchestrator retrieves the relevant data and invokes the visualization agent. The agent interprets the query and data schema to generate several visualizations capturing various aspects of system health. Figure~\ref{fig:calls-distribution} displays one of the resulting charts. It shows the temporal distribution of application calls, aiding SREs in verifying whether the system exhibits any anomalous behavior. In this case, the call pattern remains within expected bounds over the whole window.

\subsubsection*{Key Observations}

Restricting the agent to five chart types limits hallucinated encodings, but it also causes most legality failures on specific queries. Supporting more chart types, or telling the user when a requested type is replaced, would fix this. During development, prompts that included detailed statistical summaries tended to give worse charts. Row-level preprocessing is not supported, because whole datasets exceed the LLM's input length limit.

\section{Discussion}
\label{sec:discussion}

\subsection*{Engineering Lessons}

An earlier evaluation of this system counted HTTP-level successes and used a hand-tuned quality score. Such measures show whether agents respond, not whether they are right, so every result in this paper is scored against gold answers or, for schema mapping and chart quality, a model judge. Scoring against answers is also what exposed the defects reported here, since an agent that returns no chart or calls the wrong tool still responds successfully at the HTTP level. The choice of metric matters as much as the choice of answers. Routing accuracy above 0.9 hides that the keyword router almost never routes GraphQL query requests correctly, because almost all requests are schema mapping requests that a single keyword catches. A router that ignored its input and always chose schema mapping would still score 0.82 accuracy, against a macro-F1 of 0.23. Macro-F1 and per-class recall reveal this, and they also show that rules that override the LLM can lower recall on the minority GraphQL class. The same caution applies to reference-based scores for generated artifacts. The schema generation agent agrees with the hand-written schemas on fields far more than on types. Those schemas are one curated design, not the only correct answer. A functional test would measure correctness more directly than agreement does, for example by checking whether the queries SREs need can be answered over the generated schema.

Model judges deserve the same scrutiny as the agents they score. Two of our results, schema mapping and chart quality, rest on a judge rather than on gold answers, and the judge's own criteria shape the numbers. The chart judge marks a substituted chart type as illegal, so its legality score measures how closely a chart follows the request rather than how useful the chart is. The original system was judged by Pixtral-12B and the re-run by Mistral Medium, which is one reason the two sets of chart scores cannot be compared directly. Calibrating each judge against human labels on a sample of its inputs, and reporting the agreement, would turn these scores from estimates into measurements. The same labels would also serve as gold answers for later versions of the agents.

Two agents had defects that appeared only when we measured them with current models. The visualization agent's prompt and parser disagreed about Markdown code fences. The prompt told the model not to use them while the parser required them, so the deployed pipeline produced no charts. A contract test, which runs the parser on real model output, would have caught this. The DLM agent depended on the serving stack, the gateway and inference server that run the model. The same Llama model ignored tool results behind one gateway and not another, because of how its chat template formats system messages into the model's input. Both defects sit between the model and the code around it, where unit tests with mocked model output pass by construction. Cemri et al.~\cite{cemri2025multiagentfail} group the failures of multi-agent LLM systems into specification, inter-agent misalignment and task verification. The two defects here lie below these, between a single agent and the stack that serves its model. Testing on the target serving stack, with the prompts, parsers and chat templates that production uses, is part of testing an agent. It must be repeated when the model or gateway changes. Agent analytics supports this diagnosis. Its traces connect a wrong answer to the routing decision, tool call or model call behind it, and attribute latency and model usage to each agent.

Re-running the system a year after it was built also tested the architecture in a way we had not planned. The Granite-3.x models that the agents were built on are no longer served, so every agent had to move to a new model. Each change this needed, described in the agent's own section, stayed inside that agent, which is the isolation the three-tier design aims for. The move also showed that model retirement is an operational risk for agent systems. Prompts tuned for one model family can fail silently with another, as the code-fence defect did. Pinning model versions, keeping the evaluation harness runnable and re-running it whenever a model is replaced matter as much as the agents themselves.

\subsection*{Design Implications}

The routing results suggest how rules and an LLM should share the work in an orchestrator. Keyword rules are cheap and predictable, and they catch almost every schema mapping request, since nearly all start with the same verb. They fail on requests whose wording overlaps another agent's, as with GraphQL query requests phrased as display requests. In the hybrid router, a firing rule then overrides an LLM that would have routed the request correctly. Rules are therefore best kept for high-precision patterns, with every other request left to the LLM, or passed to the LLM as hints rather than decisions. Because the agent cards already describe each agent's skills, the LLM router needs no new configuration when an agent is added, while a rule-based router needs rules for each agent.

The agents also show where data should and should not enter the model's context. The visualization agent gives the LLM the schema of the data, with column names and types, rather than the rows, and row-level preprocessing is left out because of input length limits. Detailed statistical summaries in the prompt tended to make its output worse. The GraphQL layer fits this pattern, since the query agent reasons over the schema and the query retrieves only the fields a request needs, as in the example of Section~\ref{sec:graphql-query}. Keeping rows out of the context and passing the model schemas, queries and small results is likely to matter more as observability data grow, since metrics and traces quickly exceed any context window.

\subsection*{Limitations}

The data limit how far these results generalize. The DLM scenarios and their backends are synthetic, the visualization data are small samples, and the routing labels come from a dataset generator rather than human annotators. The requests themselves come from the same generator, so their phrasing may be more regular than that of real operators. Several per-class and per-type samples are small, and each result is one greedy run. IBM watsonx outputs are not bit-reproducible, so the same request can return slightly different answers, and small gaps deserve caution. Schema mapping is judged by an LLM without gold answers, and the GraphQL benchmark uses general-domain rather than observability schemas. The original agents used Granite-3.x models no longer served, and we evaluate with Llama-3.3-70B and Granite-4-H-Small, so results for the original configuration are not reproduced. Finally, we evaluate agents individually. End-to-end workflows, the overhead of the A2A and MCP layers, the scalability of a single orchestrator to many agents, and agent behavior when external services fail are not measured here. The DLM results show that the agent reasons and calls its tools correctly against deterministic mocks, not that it works with live Kubecost and live tracing systems.

\section{Conclusion}

We presented a three-tier reference architecture for coordinating data workflows in enterprise IT, which combines specialized LangGraph agents, MCP tool integration and A2A inter-agent communication, and is instrumented with agent analytics. Its data access layer generates GraphQL queries and schemas and maps between product schemas, and its consumption agents build dashboards and manage trace retention. The agents share open protocols, so each can be replaced on its own. The evaluation shows that query generation on a public benchmark and routing work well with a current open model, and that the data lifecycle management agent solves most of its scenarios. Schema generation agrees with human designs on fields more than on types, and two agents had deployment defects that were found only when measured. The fixes are to test prompts and parsers together, test on the target serving stack, and report macro metrics. Next we will apply these fixes, evaluate end-to-end workflows through the agent analytics traces, and adapt models to our schemas with parameter-efficient fine-tuning.

\clearpage
\bibliographystyle{ACM-Reference-Format}
\bibliography{icse-2026}

\end{document}